\pdfoutput=1

\documentclass[11pt]{article}
\usepackage{acl}

\usepackage{times}
\usepackage{latexsym}
\usepackage{graphicx}
\usepackage{amsfonts}
\usepackage{amsmath} 
\usepackage{booktabs}
\usepackage{hyperref}
\usepackage{float} 
\usepackage{multirow}
\usepackage[T1]{fontenc}

\usepackage[utf8]{inputenc}

\usepackage{microtype}

\usepackage{inconsolata}

\title{OMEGA: Optimized Multimodal Position Encoding Index Derivation with Global Adaptive Scaling for Vision-Language Models}

\author{Ruoxiang Huang, Xindian Ma, Rundong Kong, Zhen Yuan, Peng Zhang\thanks{Corresponding author.}\\
 Tianjin University, Tianjin, China \\
  \texttt{\{huangruoxiang\_0927,xindianma,krd,zhenyuan\_1219,pzhang\}@tju.edu.cn} 
  }

\begin{document}
\maketitle
\begin{abstract}
Vision-Language Models (VLMs) have demonstrated strong performance across various multimodal tasks, where position encoding plays a vital role in modeling both the sequential structure of textual information and the spatial structure of visual information. However, current VLMs commonly adopt modality-unified 1D or 2D positional indexing strategies, which treat textual and visual tokens uniformly without accounting for their distinct structural properties and sequential continuity for text and spatial coherence for vision. To address this limitation, we propose OMEGA, a novel position encoding framework that employs Modality-Specific Position Encoding (MSPE) to assign positional indices while preserving the inherent structures of each modality across separate coordinate dimensions. Additionally, to align the information density of multimodal data in the positional index space, OMEGA introduces Global Adaptive Encoding Step Scaling (GAESS), which adaptively adjusts the position encoding step size of visual tokens based on the embedding entropy of both modalities. Experimental results demonstrate that OMEGA consistently enhances VLM performance across diverse architectures and VQA benchmarks. On visual-intensive tasks, OMEGA achieves up to 3.43\% improvement over baseline position encoding strategies on Qwen2.5-VL-3B, with consistent gains observed across larger models including Qwen2.5-VL-7B and LLaVA-v1.5-7B.

\end{abstract}

\section{Introduction}

In recent years, with the rapid advancement of Large Language Models (LLMs) \cite{GPT3,PaLM} and Vision-Language Models (VLMs) \citep{yang2023dawnlmmspreliminaryexplorations,bai2025qwen25vltechnicalreport,chen2025janusprounifiedmultimodalunderstanding} multimodal artificial intelligence systems have achieved significant progress in tasks such as visual question answering, image captioning and complex visual reasoning. 

Recent studies \citep{Dhamyal-CapturingModality-2022,ge2024v2peimprovingmultimodallongcontext} have demonstrated that Position Encoding (PE) plays a crucial role in VLMs. The positional index derivation strategy forms the foundation of PE and plays a crucial role in enabling model to understand the sequential and spatial relationships of textual and visual information. 
Existing VLMs typically adopt a Modality-Unified position encoding strategy with fixed encoding step size, applying encoding schemes designed for pure textual data to multimodal inputs. 
Specifically, there are two types of positional index derivation strategies: 1-Dimension PE(1D-PE)
and 2-Dimension PE (2D-PE).

\begin{figure*}[ht]
\centering  \includegraphics[width=1.0\textwidth]{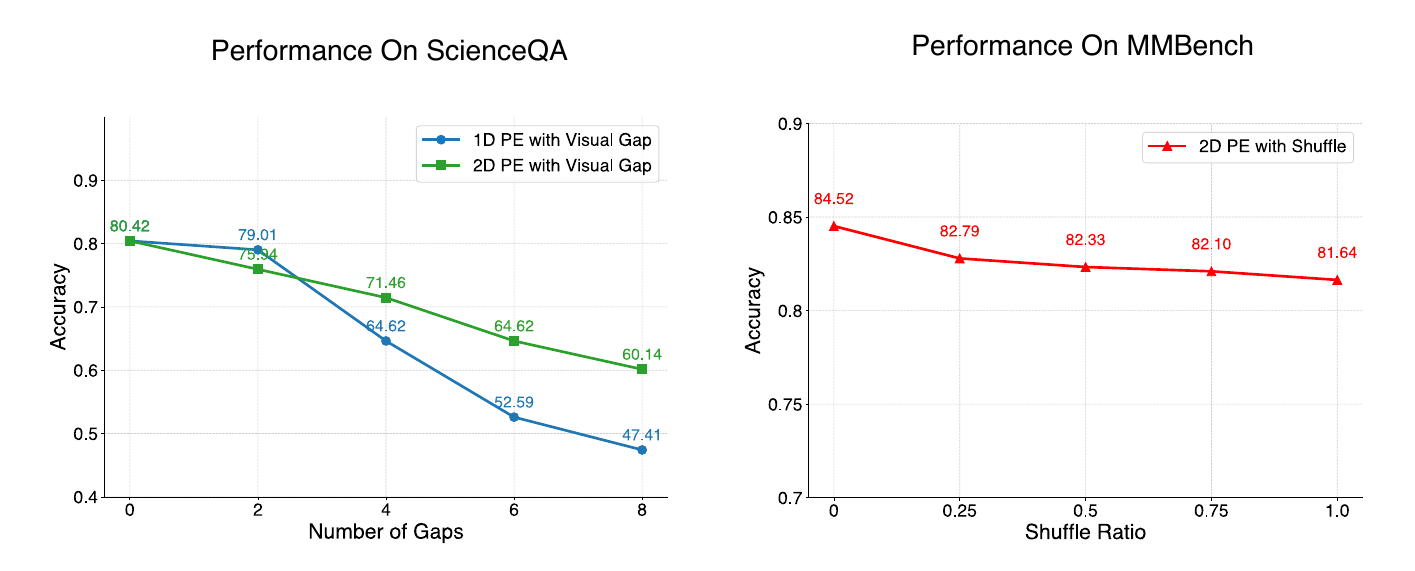}
  \caption{Empirical analysis of sequential continuity and spatial structure disruption. Sequential continuity  is disrupted by randomly inserting positional index gaps into the text sequence, with each gap sized to match the number of visual tokens per image. Spatial structure is disrupted by randomly shuffling the positional indices of a proportion of visual tokens during index derivation. Left: The relationship between the number of Visual Gaps and the accuracy of QwenVL2.5-VL-3B on ScienceQA. Right: The relationship between the proportion of shuffled visual tokens and the accuracy of QwenVL2.5-VL-3B on MMBench.}
  \label{fig:empirical analysis}
\end{figure*}

Under 1D-PE and 2D-PE strategies, a substantial number of visual tokens are inserted into the textual token sequence, disrupting the contextual continuity of textual information in the positional index space. 
As shown in Figure \ref{fig:empirical analysis}, experimental results demonstrate that model performance declines as the number of visual gaps increases, indicating that the disruption of positional continuity in the textual token sequence caused by the insertion of visual tokens negatively impacts the model's performance. Specifically, for more general 1D-PE strategy, visual and textual tokens are arranged in a unified sequential order, resulting in the disruption of the spatial structure of visual information. 

Experimental results indicate that the model’s performance declines as the proportion of shuffled visual tokens increases, confirming the negative impact of disrupted spatial structure on model performance.

To address the above issues, we propose a novel PE framework \textbf{OMEGA}, which consists of two components, \textbf{Modality-Specific Position Encoding (MSPE)} and \textbf{Global Adaptive Encoding Step Scaling (GAESS)}. MSPE assigns separate positional index dimensions to textual and visual tokens, introduces placeholders in the text sequence to indicate the positions of visual tokens, and preserves both sequential continuity and spatial structure. Inspired by prior studies\cite{palepu2023tiertextimageentropyregularization, denseinformation}, we consider that aligning the information density between textual and visual modalities can enhance model’s textual-visual understanding capability. GAESS achieves this alignment by computing the embedding entropy of textual and visual tokens and adaptively scaling encoding step size of visual tokens to align their information density in the index space.

In our experiments, we applied OMEGA to enhance the textual-visual understanding capabilities of the open-source VLM Qwen2.5-VL-3B. The results demonstrate that our method improves performance on Visual Question Answering (VQA) tasks. OMEGA achieves strong results across multiple VQA benchmarks. Compared to the model's original 2D-PE, OMEGA delivers average improvements of 1.87\% in the zero-shot setting and 0.84\% in the fine-tuned setting. Such performance gains are notable for an approach that solely modifies the positional index derivation without altering the model architecture or attention mechanism.

The contributions of this work are as follows: 1) We propose the OMEGA position encoding framework and demonstrate its effectiveness in improving VLM performance on VQA tasks. 2) We introduce MSPE, which preserves both the continuity of positional indices and the spatial structure of visual information. 3) We introduce GAESS, which aligns the information density across modalities within the positional index space.

\section{Related Work}
\textbf{Vision-Language Models.} With the rise of Large Language Models (LLMs), Vision-Language Models (VLMs) have become a core paradigm in multimodal AI, aiming to integrate visual and linguistic inputs for tasks involving images and text. CLIP \citep{CLIP} leverages contrastive learning to align image-text pairs. Gemini \citep{gemini} extends multimodal processing to text, images, audio, and video. LLaVA \citep{llaVA} integrates a visual encoder with an LLM and is trained end-to-end using vision-language instruction data. Several studies have focused on enhancing spatial relationship modeling in VLMs: SpatialVLM \citep{spatialvlm} projects 2D images into metrically scaled 3D point clouds, while AdaptVis \citep{adaptvis} dynamically modulates attention based on confidence estimates. The above methods require changes to the model architecture or attention mechanism. In comparison, OMEGA only modifies the position encoding, making it a low-cost solution.
\\
\textbf{Position Encoding in Transformer.} Position Encoding is a key component of the Transformer architecture \citep{vaswani-etal-2017-attention}, enabling the model to capture the sequential order of input tokens. Since the self-attention mechanism itself is invariant to input order, position encoding serves as the primary means for Transformers to incorporate positional information. Rotary Position Encoding (RoPE) \citep{su2023roformerenhancedtransformerrotary} encodes positional information by rotating embeddings in the complex plane. Its main advantage lies in its ability to naturally capture relative positional relationships while preserving absolute position information. Attention with Linear Biases (ALiBi) \citep{press2022trainshorttestlong} introduces a distance-dependent bias matrix directly into the attention computation, reducing the influence between tokens as their relative distance increases. Fourier Position Encoding (FoPE) \citep{FoPE} enhances length generalization by representing positional information as multi-frequency Fourier series. Recently, 3D Rotary Position Encoding (3D-RPE) \citep{ma20253drpeenhancinglongcontext} extends RoPE to a three-dimensional sphere, offering controllable long-term decay and improved position resolution for long-context modeling. However, these position encoding strategies do not introduce new positional index derivation strategies. Most VLMs employing them still rely on 1D positional indices. Qwen2.5-VL\citep{bai2025qwen25vltechnicalreport} adopts M-RoPE to extend positional indices into higher-dimensional spaces, enabling the encoding of both temporal and spatial information to accommodate image and video modalities. For this reason, we argue that the experimental comparisons only need to include 1D-PE and 2D-PE, rather than specific position encoding computation strategies, such as RoPE and 3D-RPE.

\begin{figure*}[htbp]
\centering
  \includegraphics[width=0.95\textwidth]{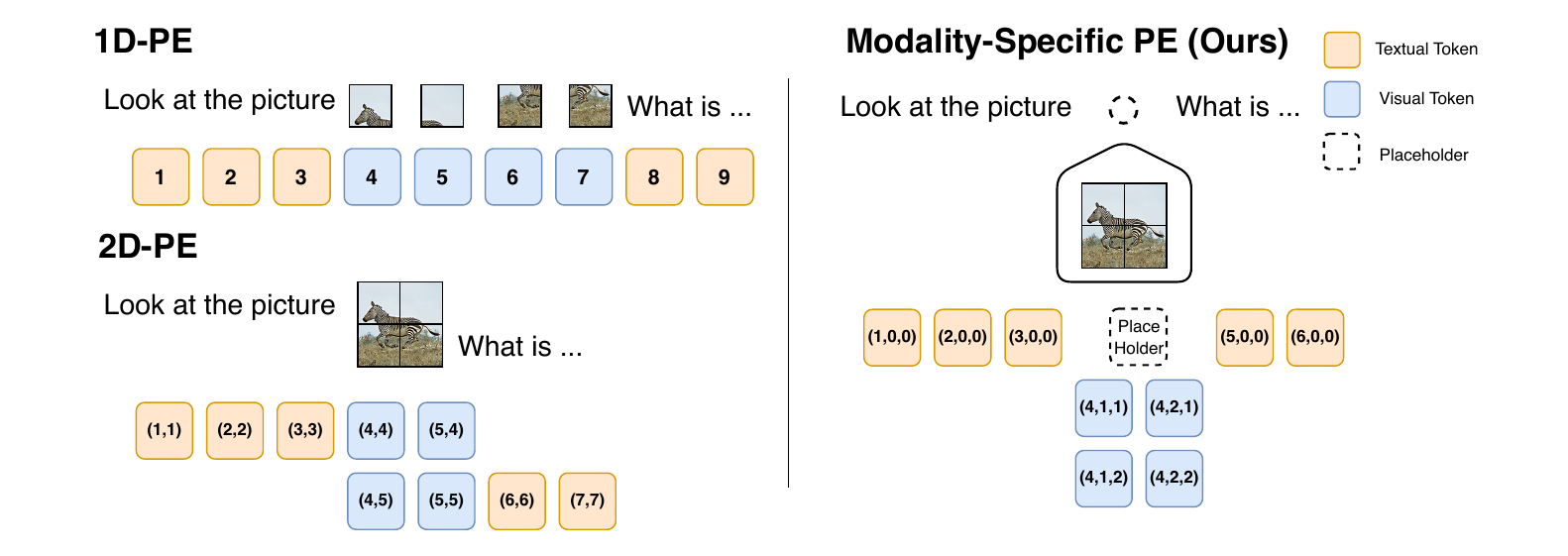}
  \caption{Illustration of Modality-Specific Position Encoding (MSPE) compared with Modality-Unified 1D-PE and 2D-PE.}
  \label{fig;MSPE}
\end{figure*}

\section{Methodology}
\subsection{Positional Index Derivation}
Position Encoding can be formalized as consisting of two steps: positional index derivation and positional embedding computation. The former assigns a positional index $p_i $ to each token $t_i$, while the latter computes the corresponding positional embedding $e_i$ based on $p_i$ and incorporates it into $t_i$ to inject positional information into the token. OMEGA focuses solely on the positional index derivation step and can be combined with any positional embedding computation method, reflecting the generalizability of our approach.

The original input to the VLM is mapped by the embedding layer into a sequence of tokens. Let $M = [t_0, t_1, \dots, t_n]$ be a sequence of multimodal input tokens, where each token $t_i \in M$ belongs to one of two disjoint sets: $\mathcal{T}$ and $\mathcal{V}$, which denote the set of textual tokens and the set of visual tokens respectively. The positional index derivation function $p$ can be formally defined as:

\begin{equation}
p: M \rightarrow \mathbb{N}^{dim}
\end{equation}
where $dim$ is the dimensionality of the positional index, determined by the position encoding strategy.

\subsection{Modality-Unified Position Encoding}
\label{sec:modality-unified}
Existing position encoding strategies in most VLMs are Modality-Unified: they assign positional indices to tokens based solely on their position within the unified token sequence, without regard to whether it is a textual or visual token. This unified treatment neglects the fundamental differences in spatial structure and semantic continuity between modalities. In what follows, we formalize two representative positional indexing schemes—1D and 2D position encodings and analyze their inherent limitations.

Under 1D-PE, all tokens share a unified linear index space, regardless of modality. The positional index derivation function is defined as:

\begin{equation}
p^{1\mathrm{D}}(t_i) = i
\end{equation}

As shown in Figure \ref{fig;MSPE}, 1D-PE neglects spatial structure of visual information. In particular, visual tokens $\mathcal{V}$, which originate from spatially structured image patches, lose their inherent 2D spatial relationships.

Under 2D-PE, each visual token $t_i \in \mathcal{V}$ corresponds to a specific image patch in the original visual input. Let $(u_i, v_i) \in \mathbb{N}^2$ denote the spatial coordinates of that patch within the image grid, where $u_i$ and $v_i$ represent the row and column indices, respectively. These coordinates are directly adopted as the positional indices of visual token. The positional index derivation function is defined as:
\begin{equation}
 p^{2\mathrm{D}}(t_i) =
\begin{cases}
(i, i) &  t_i \in \mathcal{T}, \\
(h_i, w_i)+ p^{2\mathrm{D}}(t_{\mathrm{prev}})  &  t_i \in \mathcal{V}.
\end{cases}
\end{equation}
where $t_{\mathrm{prev}}$ denotes the textual token that precedes the visual token $t_i$ in the sequence $M$. The addition in the above equation is performed element-wise across each dimension of the coordinate.

As shown in Figure \ref{fig;MSPE}, 2D-PE preserve the original spatial structure of visual tokens. However, interleaving the two modalities in the sequence $M$ disrupts the sequential continuity of textual information and injects heterogeneous semantics into the shared positional index space.

\subsection{Modality-Specific Position Encoding}\label{MSPE}
To preserve the sequential continuity of textual information and the spatial structure of visual information, we propose Modality-Specific Position Encoding (MSPE). This method leverages three-dimensional positional indices and assigns distinct coordinate 
dimensions to textual and visual tokens. We define a transformation function:
\begin{equation}
\phi: M \rightarrow M' = [t'_0, t'_1, \dots, t'_m] \quad t'_i \in \mathcal{T} \cup \mathcal{H}
\end{equation}
which replaces each visual token in the original sequence with a placeholder token while leaving textual tokens unchanged. The resulting sequence $M'$ consists only of textual and placeholder tokens, maintaining the sequential continuity of the input.

Each placeholder token $h_j \in \mathcal{H}$ is associated with a group of visual tokens via a mapping:
\begin{equation}
\psi: \mathcal{H} \rightarrow \mathcal{P}(\mathcal{V})
\end{equation}
where $\mathcal{P}(\mathcal{V})$ denotes the power set of $\mathcal{V}$. Specifically, $\psi(h_j) = \{v_{j,1}, v_{j,2}, \dots, v_{j,k}\}$ represents the set of visual tokens corresponding to placeholder $h_j$, with the constraint that all $v_{j,l}$ belong to the same image instance and are spatially located within a common image grid. Each $v_{j,l} \in \mathcal{V}$ is associated with a spatial coordinate $(h_{j,l}, w_{j,l}) \in \mathbb{N}^2$ in the original image.

We define MSPE positional index  derivation function as follows:
\begin{equation}
p^{\text{MS}}(t_i) =
\begin{cases}
(i, 0, 0) & t_i \in \mathcal{T} \cup \mathcal{H} \\
(0, h_{j,l}, w_{j,l})+p^{\text{MS}}(h_j) & t_i \in \mathcal{V}\ \\ 
\end{cases}
\end{equation}
where $h_j$ denotes the placeholder corresponding to the image instance to which visual token $t_i = v_{j,l}$ belongs, i.e., $t_i \in \psi(h_j)$. The addition in the above equation is performed element-wise across each dimension of the coordinate.

MSPE offers two key advantages for VLMs in modeling positional information within multimodal input sequences.

First, as shown in Figure \ref{fig;MSPE}, it preserves the sequential continuity of textual information and the spatial structure of visual information simultaneously. Since visual tokens are not directly interleaved into the textual token sequence, the continuity of the textual tokens remains intact. Placeholder tokens maintain positional alignment in the sequence without disrupting textual continuity, allowing the model to attend over text as a coherent 1D sequence while still accessing visual context through the shared position space.

Second, MSPE assigns distinct coordinate dimensions to textual and visual tokens, enabling dimension-wise control over positional scaling. In contrast to Modality-Unified PE, where all modalities share the same positional index dimensions, this separation allows us to apply scaling transformations to specific modalities without explicitly tagging token types. This property is particularly beneficial for GAESS, as discussed in Section \ref{sec:GAESS}.

\subsection{Global Adaptive Encoding Step Scaling}
\label{sec:GAESS}
\begin{figure}[ht]
\centering
  \includegraphics[width=\columnwidth]{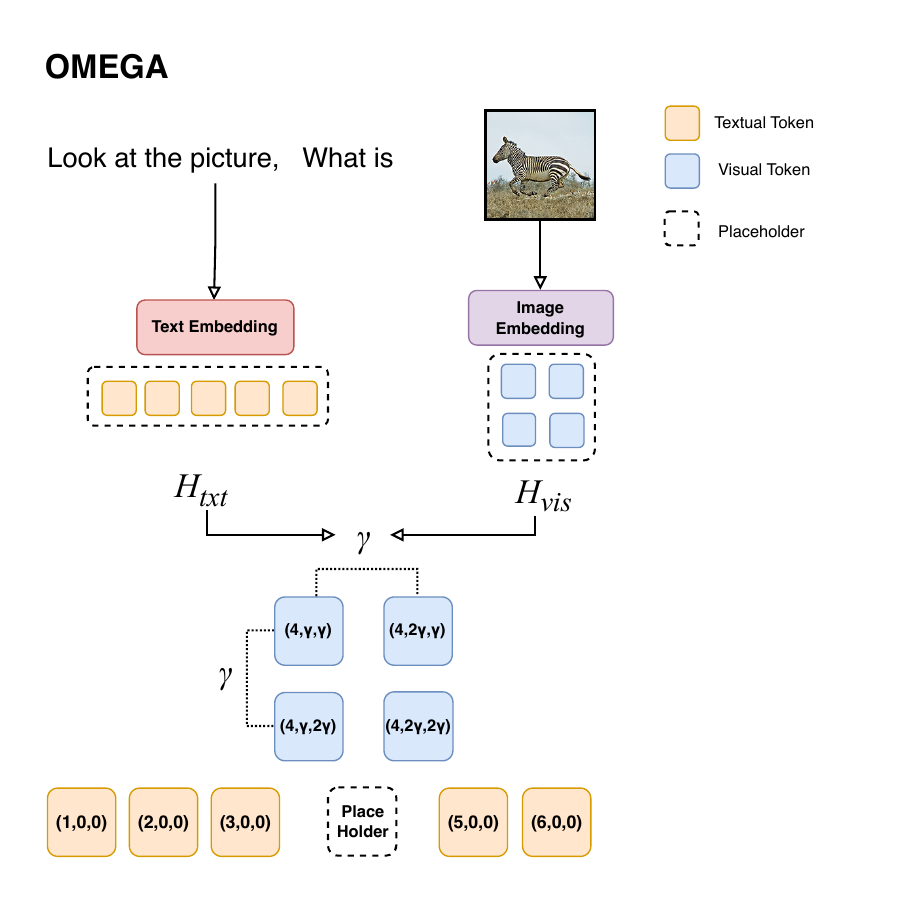}
  \caption{Illustration of Global Adaptive Encoding Step Scaling (GAESS) under Modality-Specific Position Encoding}
  \label{fig:GAESS}
  \vspace{-0.5cm}
\end{figure}

Prior work has explored the relationship between modality-specific information distribution and representation alignment. For instance, Entropy-Based Regularization was proposed to bridge text-image semantic gaps, highlighting the role of information density in enhancing contrastive vision-language learning.\citep{palepu2023tiertextimageentropyregularization} Similarly, DCLIP\cite{denseinformation} enhances the information density of textual inputs to match that of visual content, thereby improving multimodal alignment and reasoning capabilities. These findings support our hypothesis that aligning positional information density across modalities facilitates more coherent cross-modal alignment in the positional index space.

In multimodal representation learning, textual and visual modalities differ not only in structural dimensionality but also in information density. A fixed encoding step size often leads to a mismatch in positional resolution between modalities. To address this, we propose Global Adaptive Encoding Step Scaling (GAESS) that adjusts positional granularity based on the estimated information content of each modality.

Information entropy provides a simple measure of the information content encoded in feature representations. It quantifies the uncertainty or dispersion of embedding vectors, thereby reflecting how densely packed or diverse the encoded semantic patterns are. Entropy can be directly computed from raw embeddings without the need for supervision or text-image alignment.

To quantify the information content of textual and visual modalities in multimodal inputs, we propose a histogram-based marginal entropy estimation method, referred to as \textbf{Entropy Embedding}.

Given an embedding matrix \( Z \in \mathbb{R}^{N \times d} \), where \( N \) denotes the number of tokens and \( d \) the embedding dimension, we begin by constructing a histogram for each embedding dimension independently. For each dimension \( j \in \{1, \dots, d\} \), let \( \{z_{1j}, z_{2j}, \dots, z_{Nj}\} \) denote the set of values across all tokens in that dimension. For each dimension $j$, we determine the minimum and maximum values among all tokens as $z_j^{\text{min}}$ and  $z_j^{\text{max}}$ respectively. The number of histogram bins, denoted as $K$, is treated as a predefined hyperparameter. 

We partition the range \([z_j^{\text{min}}, z_j^{\text{max}}]\) into \( K \) uniform bins and denote $R_j= z_j^{\text{max}}-z_j^{\text{min}} $ .The $k$-th bin of $j$-th dimension $\mathcal{B}_j^{(k)}$ is defined as:

\begin{equation}
    \mathcal{B}_j^{(k)} = \left[ z_j^{\text{min}} + \frac{k-1}{K}R_j, \; z_j^{\text{min}} + \frac{k}{K}R_j \right] 
\end{equation}

Let \( h_{j,k} \) denote the count of embedding values \( z_{ij} \) falling into bin \( \mathcal{B}_j^{(k)} \), and define the normalized bin probability $ p_{j,k}$ as:

\begin{equation}
     p_{j,k} = \frac{h_{j,k}}{\sum_{k'=1}^{K} h_{j,k'}} 
\end{equation}

Based on these per-dimension distributions, we define the Embedding Entropy of the modality as the average marginal entropy across all dimensions:

\begin{equation}
H(Z) = \frac{1}{d} \sum_{j=1}^{d} \left( - \sum_{k=1}^{K} p_{j,k} \log_2 p_{j,k} \right)
\label{eq:entropy_embedding}
\end{equation}

Embedding Entropy is measured in bits per dimension, reflects the average uncertainty or dispersion of embedding values. We denote the embedding entropy of textual and visual modalities as $H_{\text{txt}}$ and $H_{\text{vis}}$, respectively. This method is computationally efficient and does not require supervision or alignment pairs.

To ensure consistent information densities across modalities in the positional index space, we aim to align the effective information per positional unit. Specifically, we define the \textbf{Information Density Ratio} in positional index space $\rho$ as:

\begin{equation}
    \rho = \sqrt{\frac{H_{\text{vis}}}{H_{\text{txt}}}}
\end{equation}

The definition is derived from the requirement to align the positional information density between textual and visual modalities under MSPE. In this setting, textual tokens are arranged along a 1D positional axis, while visual tokens are distributed over a 2D spatial grid. To ensure that each modality contributes equivalent information per unit positional region, we compare the textual information per unit length with the visual information per unit area in vision. In practice, we set the encoding step size of textual tokens to a fixed value of $1$ and we aim to satisfy: $\frac{H_{\text{vis}}}{\gamma^2} = H_{\text{txt}}$, where $\gamma$ is the step size applied to both spatial dimensions of visual tokens. Accordingly, $\gamma$ acts as global visual scaling factor that adjusts the positional resolution of visual tokens to achieve modality-aligned information density in the index space.

However, in practice, embedding entropy estimates may vary significantly across samples and domains, potentially leading to unstable scaling behaviors. To prevent extreme positional resolutions, we introduce lower and upper bounds, $\gamma_{\text{min}}$ and $\gamma_{\text{max}}$, on the scaling factor:

\begin{equation}
    \gamma=
    \begin{cases}
    \gamma_{\text{min}} & \rho \leq \gamma_{\text{min}}\\
    \rho                & \gamma_{\text{min}} \leq \rho \leq \gamma_{\text{max}}\\
    \gamma_{\text{max}} & \rho \geq \gamma_{\text{max}}
    \end{cases}
\end{equation}

Figure \ref{fig:GAESS} illustrates the application of GAESS in Modality-Specific Position Encoding. We now present the unified formulation of \textbf{OMEGA}, which integrates MSPE with GAESS to derive the final positional index for multimodal input sequences:

\begin{equation}
p^{\mathbf{\Omega}}(t_i) =
\begin{cases}
(i, 0, 0), ~~~~~~~~~~~~~~~~~~~~~~~~ t_i \in \mathcal{T} \cup \mathcal{H} \\
(0, \gamma h_{j,l}, \gamma w_{j,l})+p^{\mathbf{\Omega}}(h_j),~~t_i \in \mathcal{V}\ \\ 
\end{cases}
\end{equation}
This unified design simultaneously accounts for the structural heterogeneity and information density differences between textual and visual information. Appendix \ref{sec:appendix-examples} presents examples of computing embedding entropy and $\gamma$ for visual question answering problems across different datasets.

\begin{table*}[ht]
\centering
\begin{small}
\begin{tabular}{lcccccc}
\toprule
\textbf{Model} & \textbf{Position Encoding} & $\textbf{ScienceQA}_{\text{ovr}}$ & $\textbf{ScienceQA}_{\text{vis}}$ & \textbf{RealWorldQA} & \textbf{MathVision} &  \textbf{MMBench}  \\
\midrule
\multirow{5}{*}{\textbf{Qwen2.5-VL-3B}} 
& No PE                         & 39.62 & 33.33 & 33.33 & 14.80 & 5.89 \\
& MIPE                          & 77.12 & 71.08 & 63.23 & 21.52 & 83.69 \\
& 1D-PE                         & 73.35 & 77.94 & 60.66 & 22.04 & 84.20 \\
& 2D-PE                         & 80.42 & 78.92 & 63.23 & 21.71 & 84.52 \\
\cmidrule{2-7}
& OMEGA                         & \textbf{82.31} & \textbf{82.35} & \textbf{64.17} & \textbf{22.70} & \textbf{85.72} \\
\midrule
\multirow{5}{*}{\textbf{Qwen2.5-VL-7B}} 
& 1D-PE                         & 84.91 & 84.80 & 67.58 & 22.77 & 87.18 \\
& 2D-PE                         & 81.84 & 84.80 & 67.81 & 23.43 & 87.25 \\
& V2PE                          & 81.37 & 84.31 & 67.58 & 23.10 & 87.18 \\
\cmidrule{2-7}
& OMEGA                         & \textbf{85.61} & \textbf{86.27} & \textbf{68.72} & \textbf{25.41} & \textbf{87.39} \\
\midrule
\multirow{5}{*}{\textbf{LLaVA-v1.5-7B}} 
& 1D-PE                         & 69.58 & 67.65 & 44.52 & 16.78 & 72.70 \\
& 2D-PE                         & 70.28 & 68.14 & 43.61 & 15.46 & 72.82 \\
& V2PE                          & \textbf{71.46} & 68.63 & 43.38 & 15.79 & 73.04 \\
\cmidrule{2-7}
& OMEGA                         & 70.75 & \textbf{71.57} & \textbf{45.43} & \textbf{17.11} & \textbf{73.09} \\
\bottomrule
\end{tabular}
\end{small}
\caption{Zero-shot accuracy comparison across three vision-language model backbones using different position encoding strategies. OMEGA consistently achieves the best or near-best performance across diverse architectures and parameter scales.}
\label{tab:zero-shot}
\end{table*}



\section{Experiment}
To validate the effectiveness of OMEGA in enhancing VLM spatial understanding of multimodal sequences, we conduct comprehensive evaluations on a suite of Visual Question Answering (VQA) datasets. Our experiments are designed to answer three central questions: (1) Can OMEGA improve zero-shot reasoning without fine-tuning? (2) Does it maintain advantages after fine-tuning? (3) What are the respective contributions of MSPE and GAESS?

\subsection{Experimental Settings}
\subsubsection{Datasets}

We select four VQA datasets that encompass diverse forms of visual information and entail a wide spectrum of reasoning abilities, including visual inference, text recognition, and object grounding, allowing us to  assess the impact of OMEGA and other position encoding strategies on the textual-visual understanding capabilities of VLM comprehensively.\\
\textbf{ScienceQA} \citep{lu2022learn}: a large-scale multimodal science question answering dataset designed to evaluate multimodal reasoning capabilities covering various forms of information including text, images and tables.\\ 
\textbf{RealWorldQA} \citep{realworldqa2024}: a VQA dataset designed to evaluate the spatial understanding capabilities of multimodal models in real-world settings. \\ 
\textbf{MathVision} \citep{wang2024measuring}: a VQA dataset specifically designed to evaluate the performance of multimodal large models on mathematical reasoning tasks. \\
\textbf{MMBench} \citep{liu2024mmbench}: a VQA benchmark dataset specifically designed to systematically evaluate the comprehensive capabilities of multimodal large models across perceptual and cognitive tasks. \\
For the source and detailed information of the dataset, please refer to Appendix \ref{sec:appendix-datasets}.

\subsubsection{Backbones and Position Encoding Variants}
To assess the generalizability and robustness of OMEGA across diverse backbones and parameter scales, we conduct experiments on three representative vision-language models: Qwen2.5-VL-3B, Qwen2.5-VL-7B \citep{bai2025qwen25vltechnicalreport}, and LLaVA-v1.5-7B \citep{llaVA}. Within the Qwen family, the 7B variant provides a direct comparison at a larger parameter scale. LLaVA-v1.5-7B offers insights into cross-architecture transferability, as it employs a frozen CLIP vision encoder connected to Vicuna via an MLP projector, contrasting with Qwen's native trainable ViT architecture. This architectural divergence enables us to evaluate whether OMEGA's benefits generalize across different vision encoding paradigms.

To systematically investigate the effect of position encoding strategies, we implement and compare the following variants across all three backbones:
\begin{itemize}
    \item \textbf{No PE}: Complete removal of position encoding to establish a lower bound.
    \item \textbf{1D-PE}: Sequential position encoding that treats all tokens uniformly.
    \item \textbf{2D-PE}: Two-dimensional position encoding that applies spatial indexing to visual patches.
    \item \textbf{MIPE}: A variant that builds upon 2D-PE by assigning separate positional index spaces to textual and visual tokens, thereby eliminating cross-modal positional relationships.
    \item \textbf{V2PE}: A recently proposed position encoding strategy that employs smaller and variable positional increments for visual tokens compared to textual tokens \citep{ge2024v2peimprovingmultimodallongcontext}. This baseline is evaluated on Qwen2.5-VL-7B and LLaVA-v1.5-7B.
    \item \textbf{OMEGA}: Our proposed method, which integrates Modality-Specific Position Encoding (MSPE) and Guided Adaptive Entropy-based Spatial Scaling (GAESS).
\end{itemize}

For a comprehensive description of implementation details and baseline configurations, please refer to Appendix \ref{sec:appendix-models} and \ref{sec:appendix-otherPE}.
\subsubsection{Training Settings}
We conduct instruction tuning on ScienceQA \citep{lu2022learn} and MMBench \citep{liu2024mmbench} using the five position encoding strategies mentioned above. During training, each model is configured with a specific position encoding strategy to ensure it can fully leverage the spatial relation information provided by specific Position Encoding strategy. For training hyperparameters, please refer to Appendix \ref{sec:appendix-fine-tuning}.
\subsection{Experiment Results}

\begin{table*}[ht]
\centering
\begin{small}
\begin{tabular}{lccccc}
\toprule
\textbf{Position Encoding} & $\textbf{ScienceQA}_{\text{ovr}}$ & $\textbf{ScienceQA}_{\text{vis}}$ & \textbf{RealWorldQA} & \textbf{MathVision} & $\textbf{MMBench}_{\text{test}}$\\
\midrule
No PE                         & 40.67 & 32.71 & 34.43 & 14.14 & 15.13 \\
Modality-Independent-PE       & 92.45 & 90.20 & 65.57 & 17.76 & 85.22 \\
1D-PE                         & 92.35 & 90.69 & 64.64 & 20.07 & 87.99 \\
2D-PE(Original)             & 94.10 & 93.14 & \textbf{66.28} & 21.71 &  87.76\\
\midrule
OMEGA                      & \textbf{94.34} & \textbf{93.63} & 65.81 & \textbf{24.01} & \textbf{89.03}\\
\bottomrule
\end{tabular}
\end{small}
\caption{Accuracy comparison of Qwen2.5-VL-3B-Instruct using variant position encoding strategies under fine-tuned setting across different VQA datasets.}
\label{tab:fine-tune}
\end{table*}

\subsubsection{Zero-Shot Results}

We begin by evaluating different PE strategies under the zero-shot setting, where VLMs rely solely on their pretrained weights without task-specific fine-tuning. This experimental design allows us to isolate the contribution of PE strategies to the model's inherent multimodal understanding capabilities. We employ accuracy as the evaluation metric. Since ScienceQA contains pure-text questions without visual information, we separately report the overall accuracy and the accuracy on visual question answering, denoted as $\textbf{ScienceQA}_{\text{ovr}}$ and $\textbf{ScienceQA}_{\text{vis}}$, respectively.

As shown in Table \ref{tab:zero-shot}, OMEGA demonstrates consistent superiority across all three model backbones and five VQA benchmarks. On Qwen2.5-VL-3B, OMEGA achieves substantial improvements over the original 2D-PE: 3.43\% on ScienceQA$_{\text{vis}}$, 0.99\% on MathVision, and 1.20\% on MMBench. These gains are particularly pronounced on visual-intensive tasks requiring fine-grained reasoning. The 3.43\% improvement on ScienceQA$_{\text{vis}}$ demonstrates OMEGA's enhanced capability in preserving visual spatial structures while maintaining textual coherence. On RealWorldQA, OMEGA achieves a 0.94\% improvement, indicating robust performance on complex visual understanding tasks.

The effectiveness of OMEGA generalizes across different parameter scales and architectures. On Qwen2.5-VL-7B, OMEGA consistently outperforms traditional PE methods and the recent V2PE approach across all benchmarks. Notably, on MathVision, OMEGA achieves 25.41\% accuracy, surpassing 2D-PE by 1.98\% and V2PE by 2.31\%, highlighting its superiority in tasks requiring simultaneous visual perception and mathematical reasoning. On LLaVA-v1.5-7B with its frozen CLIP encoder architecture, OMEGA demonstrates remarkable cross-architecture transferability. While V2PE achieves the highest overall accuracy on ScienceQA$_{\text{ovr}}$ at 71.46\%, OMEGA excels on visual-specific questions with 71.57\% compared to V2PE's 68.63\%, and maintains consistent advantages on RealWorldQA and MMBench. This pattern indicates that OMEGA's benefits are most pronounced on tasks requiring strong visual understanding, aligning with its design objective of enhancing visual-textual alignment.

In contrast, the absence of position encoding leads to catastrophic performance degradation, with near-zero accuracy on MMBench for Qwen2.5-VL-3B and substantially reduced performance across all other benchmarks. This underscores the critical importance of positional information for VLM comprehension. Similarly, MIPE underperforms compared to methods that maintain cross-modal positional relationships, demonstrating that effective multimodal understanding requires coordinated position encoding across modalities.

These results demonstrate that OMEGA not only preserves the sequential continuity of textual information and the spatial structure of visual content, but also enhances the alignment between textual and visual modalities. The consistent improvements across diverse architectures, parameter scales, and task types validate the generalizability and robustness of OMEGA's design principles.

\begin{table*}[ht]
\centering
\begin{small}
\begin{tabular}{lccccc}
\toprule
\textbf{Position Encoding} & $\textbf{ScienceQA}_{\text{ovr}}$ & $\textbf{ScienceQA}_{\text{vis}}$ & \textbf{RealWorldQA} & \textbf{MathVision} &  \textbf{MMBench}  \\
\midrule
MSPE-\textit{only}              & 81.37 & 80.39 & 63.70 & 23.03  & 84.18\\
GAESS-\textit{only}             & 81.13 & 78.43 & 62.06 & 21.05  & 85.12\\
\midrule
2D-PE(Original)                 & 80.42 & 78.92 & 63.23 & 21.71 & 84.52\\
\midrule
OMEGA                      & \textbf{82.31} & \textbf{82.35} & \textbf{64.17} & \textbf{22.70} & \textbf{85.72}\\
\bottomrule
\end{tabular}
\end{small}
\caption{Accuracy comparison of  Qwen2.5-VL-3B-Instruct using variant methods from the ablation study across different VQA datasets.}
\label{tab:ablation study}
\end{table*}

\subsubsection{Fine-Tuned Results}
To further investigate the adaptability of OMEGA, we evaluate its performance under fine-tuned setting. Unlike the zero-shot setting, fine-tuning allows the model to update its parameters based on instruction, thereby serving a dual purpose: enabling task-specific optimization while allowing the model to fully adapt to the structural characteristics introduced by each position encoding strategy. For evaluations on ScienceQA, RealWorldQA, and MathVision, we perform fine-tuning on the train set of ScienceQA. For MMBench, we split it into separate training and testing subsets, accounting for 80\% and 20\% of the original dataset, respectively, for fine-tuning and evaluation. The testing subset of MMBench is denoted as $\textbf{MMBench}_{\text{test}}$.

As shown in Table \ref{tab:fine-tune}, OMEGA achieves the highest accuracy on four out of five datasets after fine-tuning, outperforming all baseline strategies on datasets such as ScienceQA, MathVision, and MMBench. After fine-tuning, OMEGA improves MathVision accuracy by 2.30\% and MMBench by 1.27\% compared to the original 2D-PE. These results suggest that the model, once adapted to the modality-specific spatial structure introduced by OMEGA, exhibits improved textual-visual alignment and understanding capabilities under fine-tuning.

\subsubsection{Ablation Study}
To further evaluate the individual contributions of each component in OMEGA, we conduct an ablation study by separately removing MSPE and GAESS. These experiments are performed under zero-shot setting to examine their impact on VQA benchmarks. Specifically, we construct two variant methods: one that uses only MSPE without GAESS (MSPE-\textit{only}), and another one that uses original 2D-PE with GAESS (GAESS-\textit{only}).

As shown in Table\ref{tab:ablation study}, OMEGA outperforms both variants across all evaluated benchmarks, validating the complementary benefits of MSPE and GAESS. Notably, the MSPE-only variant achieves an accuracy of 82.35\% on visual questions of ScienceQA, which is already 1.96\% higher than the baseline 2D-PE, demonstrating the effectiveness of MSPE in preserving modality-specific structure: separating positional dimensions alone substantially improves the model's ability to maintain textual continuity and visual spatial structure. On MathVision, MSPE-only surpasses GAESS-only by 1.98\%, reflecting that explicit structural continuity plays a more decisive role than information density aligning on tasks requiring fine-grained reasoning. Meanwhile, the GAESS-only variant shows moderate improvements over baseline strategies, especially on MMBench (0.60\% over 2D-PE), suggesting that aligning positional index resolution with information density enhances semantic grounding even without positional index derivation changes.

These findings highlight that both MSPE and GAESS contribute independently and synergistically to the overall performance of OMEGA. Their combination consistently yields the highest accuracy, confirming that modality-specific indexing and entropy-informed scaling jointly enhance textual-visual alignment in VLMs.

\section{Conclusion}
We propose \text{OMEGA}, a novel position encoding framework that assigns different dimensions to textual and visual tokens during positional index derivation, preserving the sequential continuity of textual tokens and the spatial structure of visual tokens. Furthermore, it estimates the information density of each modality by computing the embedding entropy, and adaptively scales the global encoding step size of visual tokens to ensure comparable information density between modalities in the positional index space. Experimental results demonstrate that OMEGA consistently improves the textual-visual understanding capabilities of Vision-Language Models (VLMs) under both zero-shot and fine-tuned settings, outperforming the original and alternative position encoding strategies. These findings highlight OMEGA's potential for broader application in VLM architectures.

In the future, we plan to extend OMEGA to video-language modeling by incorporating a temporal dimension into the positional index space. This extension aims to preserve the spatial structure of visual tokens and the text sequence while introducing temporal indices for video frames. We also intend to adapt the GAESS mechanism to perform entropy-aware scaling along the temporal axis, ensuring consistent information density across spatial, temporal, and textual dimensions.

\section*{Limitations}
OMEGA is currently designed for vision-language inputs, where position encoding aligns textual sequences with spatial visual features. Its extension to other modalities such as speech, audio-visual streams, or time-series sensor data remains unexplored and may require additional adaptation to handle temporal or continuous signal structures.

Additionally, the entropy-based scaling mechanism in GAESS relies on static embedding distributions to estimate modality-specific information density. This estimation may be sensitive to the choice of embedding layer or the inherent representational characteristics of a particular model, potentially affecting the robustness of the scaling across different architectures.



\bibliography{custom}

\appendix

\section{Implement Details}
\label{sec:appendix-details}
\subsection{Datasets}
\label{sec:appendix-datasets}
The datasets employed in our experiments have all been retrieved from HuggingFace, all publicly available for research use.\\
\textbf{ScienceQA}: \href{https://huggingface.co/datasets/derek-thomas/ScienceQA}{derek-thomas/ScienceQA}, distributed under the Apache License 2.0. ScienceQA consists of 21,208 multiple-choice questions in total, among which 10,322 questions contain visual information. Each multiple-choice question includes 2 to 5 answer options, and some questions are accompanied by hints that may assist the model in selecting the correct answer. The dataset is divided into three subsets: train, validation, and test. Specifically, the train subset contains 12,726 questions, while both the validation and test subsets contain 4,241 questions each. Ground truth annotations are provided for all subsets, but only the test subset is used for evaluation in our experiments.\\
\textbf{RealworldQA}: \href{https://huggingface.co/datasets/xai-org/RealworldQA}{xai-org/RealworldQA}, distributed under the Apache License 2.0. RealWorldQA consists of 765 multiple-choice questions, each accompanied by a real-world image. Each multiple-choice question includes between 2 and 4 answer options. The dataset is not split into subsets, and all questions are annotated with ground truth answers. We use the entire dataset for evaluation.\\
\textbf{MathVision}: \href{https://huggingface.co/datasets/MathLLMs/MathVision}{MathLLMs/MathVision}, distributed under the Apache License 2.0. MathVision contains 3,040 questions in total, including 1,532 multiple-choice questions and 1,508 free-form questions. Each question is accompanied by a relevant image. Each multiple-choice question includes between 2 and 5 answer options, while each free-form question has a numeric answer or a coordinate. The dataset is not split into subsets, and all questions are annotated with ground truth answers. We use the entire dataset for evaluation.
\textbf{MMBench}: \href{https://huggingface.co/datasets/lmms-lab/MMBench}{lmms-lab/MMBench}, distributed under the Apache License 2.0. MMBench consists of 10,997 multiple-choice questions, each accompanied by a relevant image. The dataset is divided into two subsets: development and test, containing 4,330 and 6,667 questions respectively. Ground truth annotations are provided only for the development set, which we use for evaluation.\\

\subsection{Models}
\label{sec:appendix-models}
The weights for the models we have used have all been retrieved from HuggingFace. \href{https://huggingface.co/Qwen/Qwen2.5-VL-3B-Instruct}{Qwen2.5-VL-3B-Instruct}, \href{https://huggingface.co/Qwen/Qwen2.5-VL-7B-Instruct}{Qwen2.5-VL-7B-Instruct}, \href{https://huggingface.co/liuhaotian/llava-v1.5-7b}{LLavA-v1.5-7B}.

\subsection{Implementation of Other Position Encoding Strategies}
\label{sec:appendix-otherPE}
In addition to MSPE and the original 2D position encoding used in the baseline model, we implemented four additional position encoding strategies in the experiments: no position encoding, modality-independent position encoding, and 1D position encoding. The implementation details of each strategy are provided below:\\
\textbf{No Position Encoding}:We simply set all positional indices to zero to eliminate position encoding.\\
\textbf{Modality-Independent Position Encoding}:
In this strategy, textual and visual tokens are assigned positions from separate positional spaces. Specifically, textual tokens follow a continuous linear index based on their order in the sequence, while visual tokens are independently assigned 2D spatial coordinates based on their location in the image grid. The positional index function is defined as:
\begin{equation}
    p_{(\mathrm{MI})}(t_i) =
\begin{cases}
(i, i), & t_i \in \mathcal{T}, \\
(h_i, w_i), & t_i \in \mathcal{V}.
\end{cases}
\end{equation}

\begin{figure}[ht]
\centering
  \includegraphics[width=\columnwidth]{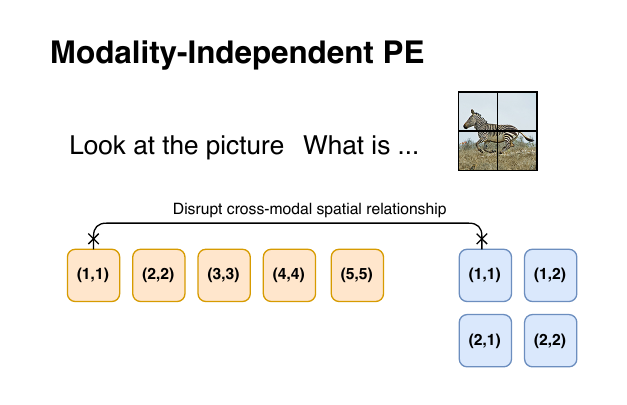}
  \caption{Illustration of Modality-Independent Position Encoding, which disrupts the spatial relationships between cross-modal tokens. }
  \label{fig:ModelIndependentPE}
\end{figure}

Figure \ref{fig:ModelIndependentPE} illustrates this strategy. As shown, it disrupts the positional continuity across modalities and weakens the model’s capacity to capture joint spatial relationships.\\
\textbf{1D Position Encoding}: Textual tokens retain their original position encoding, while visual tokens are arranged in a row-major order to form a 1D token sequence.
\textbf{V2PE}: Textual tokens retain their standard positional increments, while visual tokens are assigned smaller fractional increments (step size = 1/16 or 1/256).

\subsection{Fine-Tuning}
\label{sec:appendix-fine-tuning}
We utilized the open-source large model training framework \href{https://github.com/hiyouga/LLaMA-Factory}{LLaMA-Factory} to perform instruction tuning on Qwen2.5-VL-3B-Instruct. We modified the relevant codebase to allow the model to select different position encoding strategies during fine-tuning. The fine-tuning dataset consists of the training subset of ScienceQA, which was converted into the Alpaca format. The fine-tuning parameters are listed in Table \ref{tab:training_parameters}.

\begin{table}[htbp]
\centering
\begin{small}
\begin{tabular}{lc}
\toprule
\textbf{Training Parameter} & \textbf{Value}  \\
\midrule
Fine-Tuning Type             & LoRA     \\
LoRA Rank                    & 32  \\
LoRA $\alpha$                  & 64    \\
LoRA Target                  & all    \\
Batchsize                    & 4     \\
Gradient Accumulation Steps  & 8    \\
LR                           & $5\times10^{-5}$  \\
LR Scheduler                 & cosine \\
Epochs                       & 1  \\
Warmup Ratio                 & 0.1 \\
Precision                    & bf16 \\
\bottomrule
\end{tabular}
\end{small}
\caption{The training parameters used for supervised fine-tuning for the models with different position encoding strategies}
\label{tab:training_parameters}
\end{table}

\subsection{Prompts}
During evaluation, we designed prompts to guide the VLM in correctly understanding the questions and generating effective responses. To ensure fairness, we maintained prompt consistency across different datasets as much as possible. The following is a prompt example for multiple-choice questions:

\begin{verbatim}
Question: Which figure of speech is used 
in this text? <image1>
Options: ["euphemism","paradox"]
Please choose the correct option to answer 
the question. You must answer with exactly 
one letter: A, B, C, D or E.
\end{verbatim}

The following is a prompt example for free-form questions in MathVision:

\begin{verbatim}
Question: How many different digits can you 
find in this picture? <image1>
Please solve this math problem based on the
image and provide the numerical answer only. 
No explanations or additional text.
\end{verbatim}

\subsection{Hyperparameters}
Table \ref{tab:hyperparameters} lists the hyperparameters used in OMEGA and their corresponding values as configured in our experiments.

\begin{table}[htbp]
\centering
\begin{small}
\begin{tabular}{lcc}
\toprule
\textbf{OMEGA Hyperparameter} & \textbf{Symbol} & \textbf{Value}  \\
\midrule
Number of Histogram Bins   &$K$     & 256     \\
Minimum value of $\gamma$      &$\gamma_{\text{min}}$    &0.25\\
Maximum value of $\gamma$      &$\gamma_{\text{max}}$    & 3.0\\
\bottomrule
\end{tabular}
\end{small}
\caption{The hyperparameters used in OMEGA and their corresponding values in our implementation.}
\label{tab:hyperparameters}
\end{table}

\section{Examples of Calculating Embedding Entropy and Gamma}
\label{sec:appendix-examples}
\begin{figure*}[hbt]
\centering
  \includegraphics[width=\textwidth]{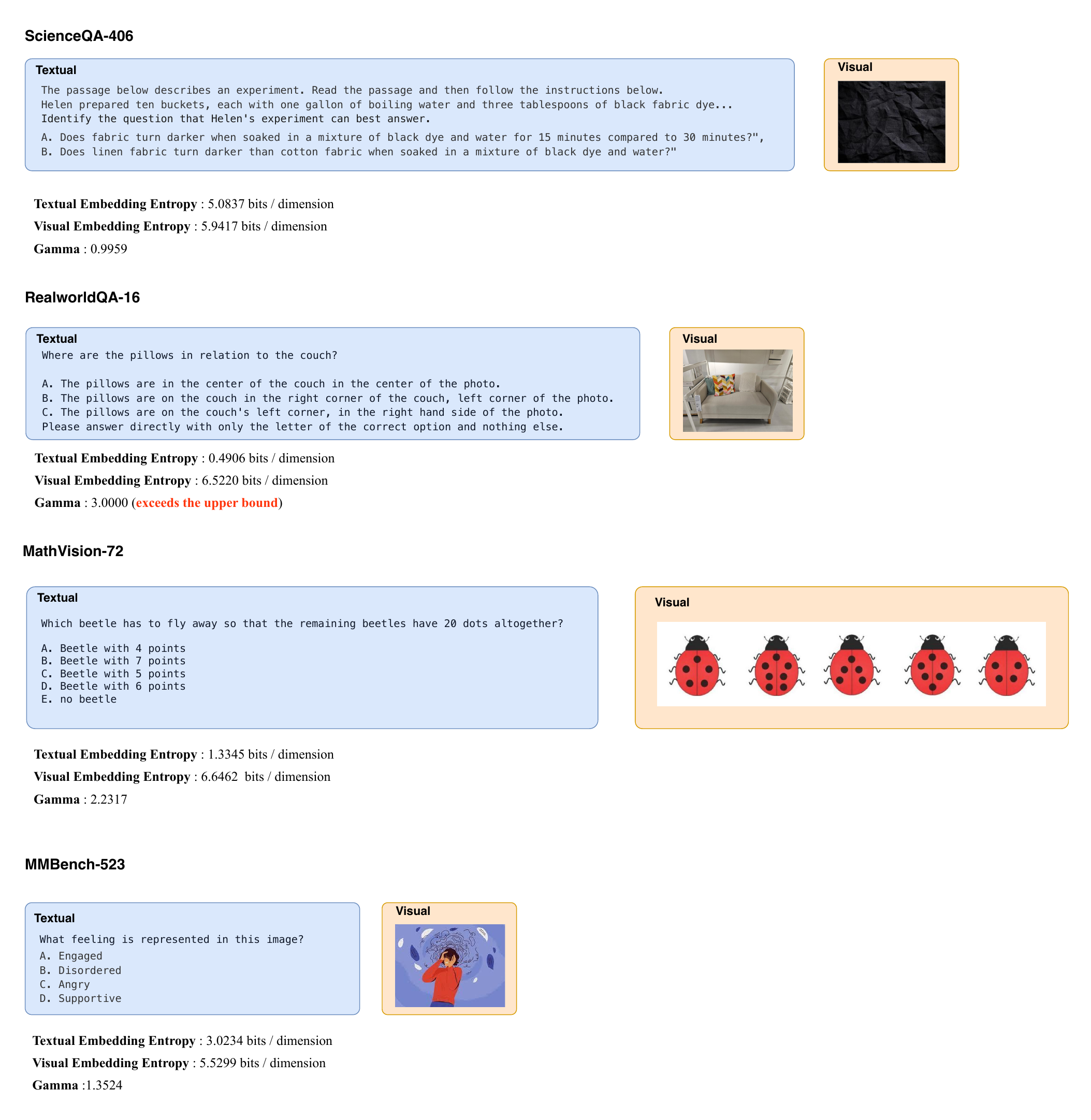}
  \caption{Examples of computing information entropy and $\gamma$ across different datasets.}
  \label{fig:case}
\end{figure*}

Figure \ref{fig:case} presents examples of GAESS calculating textual and visual embedding entropies, along with the scaling factor $\gamma$, across different datasets. These examples demonstrate the validity of using embedding entropy to quantify the amount of information conveyed by text and images. For textual information, embedding entropy exhibits high variability—texts with lower redundancy tend to have higher entropy. For visual information, the entropy differences are more moderate: images with richer details generally produce higher embedding entropy. Specifically, illustrated graphics tend to have higher entropy than uniform material textures, while real-world photographs captured by cameras typically exhibit the highest entropy among the three.

\label{sec:appendix-results}

\end{document}